# Deep Learning based Automatic Quantification of Urethral Plate Quality using the Plate Objective Scoring Tool (POST)

Tariq O. Abbas[a,b,c,d], Mohamed AbdelMoniem[e], Ibrahim Khalil[f], Md Sakib Abrar Hossain[e], Muhammad E. H. Chowdhury[e]

a. Pediatric Urology Section, Sidra Medicine, Doha, Qatar
b. College of Medicine, Qatar University, Doha, Qatar
c. Weill Cornell Medicine Qatar, Doha, Qatar
d. Regenerative Medicine Research Group, Department of Health Science and Technology, Aalborg University, Aalborg, Denmark
e. Department of Electrical Engineering, Qatar University, Doha 2713, Qatar
f. Urology Department, Hamad Medical Corporation, Doha, Qatar

**ABSTRACT**

**Objectives:** To explore the capacity of deep learning algorithm to further streamline and optimize urethral plate (UP) quality appraisal on 2D images using the plate objective scoring tool (POST), aiming to increase the objectivity and reproducibility of UP appraisal in hypospadias repair.

**Methods:** The five key POST landmarks were marked by specialists in a 691-image dataset of prepubertal boys undergoing primary hypospadias repair. This dataset was then used to develop and validate a deep learning-based landmark detection model. The proposed framework begins with glans localization and detection, where the input image is cropped using the predicted bounding box. Next, a deep convolutional neural network (CNN) architecture is used to predict the coordinates of the five POST landmarks. These predicted landmarks are then used to assess UP quality in distal hypospadias.

**Results:** The proposed model accurately localized the glans area, with a mean average precision (mAP) of 99.5% and an overall sensitivity of 99.1%. A normalized mean error (NME) of 0.07152 was achieved in predicting the coordinates of the landmarks, with a mean squared error (MSE) of 0.001 and a 20.2% failure rate at a threshold of 0.1 NME.

**Conclusions:** This deep learning application shows robustness and high precision in using POST to appraise UP quality. Further assessment using international multi-centre image-based databases is ongoing. External validation could benefit deep learning algorithms and lead to better assessments, decision-making and predictions for surgical outcomes.

## 1. INTRODUCTION

Hypospadias is one of the most common male external genital malformations and has a wide spectrum of phenotypic presentations. It is a multifactorial condition that affects the penis with a global prevalence of 3.7 per 1,000 newborns [1,2] and is predisposed by environmental and genetic factors [3]. Over 20 genes have been associated with isolated hypospadias, supporting the theory that hypospadias is a common phenotypic expression of differing genotypes [4]. These findings may partially explain the variable post-urethroplasty outcomes observed among experienced surgeons performing the same surgical techniques [4–6].

Although several important achievements have been made and numerous surgical procedures introduced in the field of hypospadiology, a significant number of post-operative complications still exist [7–9]. Several key anatomical characteristics have been linked to the occurrence of such complications and are considered important when choosing among the different surgical approaches. These factors include degree of penile curvature, urethral plate (UP) width, glans size, and meatal location [10,11].

UP preservation has been considered the landmark for several surgical techniques for hypospadias repair that function by tubularization of the UP, including the tubularized incised plate (TIP) urethroplasty, Thiersch-Duplay urethroplasty, and dorsal inlay TIP [12,13]. It is clear that UP quality could affect the procedure selection process and impact post-operative results, as we discussed in our previous systematic review [14]. However, traditional strategies for appraising UP quality are highly subjective, with a low index of generalizability. We have recently introduced the plate objective scoring tool (POST) as a reproducible and precise approach to quantifying UP quality and have published a guide to selecting particular surgical techniques based on POST [15–17]. However, there is still an inherent potential for subjectivity in the individual assessment of UP quality and POST when measuring from 2D images. This variability could cause challenges when comparing outcomes between centres and surgeons, despite attempts at standardization.

Although artificial intelligence (AI) is yet to be wholly accepted and explored in hypospadiology, it has certainly brought new possibilities to light. AI has been used in pediatric urology for predicting the outcome of surgical procedures [18], detecting the severity of the condition on the basis of imaging [19], and detecting image abnormalities [20]. A novel machine leaning (ML) model was developed to predict the future risk of febrile urinary tract infections (UTIs) related to vesicoureteral reflux [21], thus enabling personalized treatment. Recently, an ML model emulated expert human classification of patients with distal/proximal hypospadias [22].

The development of POST provides a new opportunity to apply AI to image analysis, potentially aiding surgeons in the objective selection of surgical procedures. Our goals in this study were (1) to further optimize the utilization of the POST score by using deep learning model to further streamline its function, and (2) to create a user-friendly platform to standardize POST measurement.

## 2. MATERIALS AND METHODS

Our proposed framework consists of three main stages (Figure 1). The first includes glans localization and detection: the input image is cropped using the predicted bounding box, isolating the glans (glanular area) from the image background and providing a focused region of interest within the image to improve feature extraction and landmark detection in the later stages. Next, a deep convolutional neural network (CNN) architecture is used to predict the coordinates of the five key anatomical landmarks that define the POST score [15]. Finally, these predicted landmarks are used to assess the quality of the urethral plate (UP) in distal hypospadias.

Below, we discuss the dataset used in this study, followed by a detailed description of the proposed pipeline along with the experimental setup and the evaluation matrices.

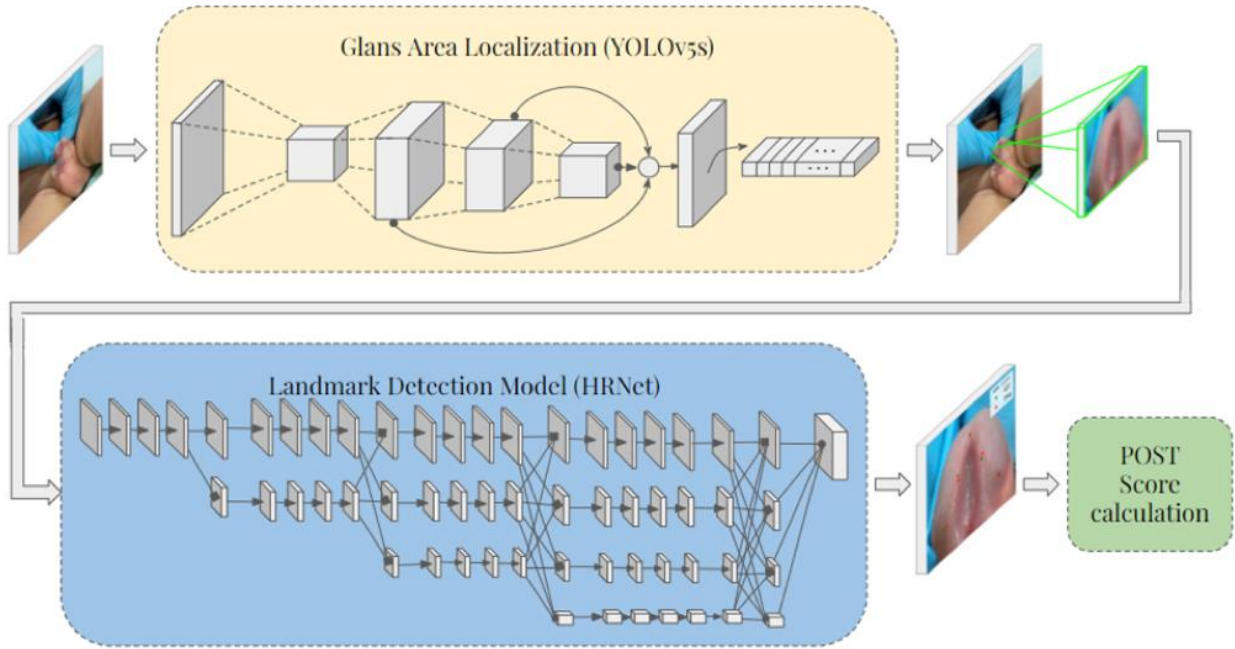

**Figure 1.** Schematic of the proposed AI-based framework, showing localization of the glans area on 2D images, automated POST landmark detection within the glans area, and POST score calculation based on these landmarks.

### 2.1 Dataset description

To train and evaluate the proposed framework, a total of 691 images were collected from prepubertal boys undergoing hypospadias repair. After IRB approval, we retrieved and used the images of hypospadias that had been gathered. Consent for the capture of these images was obtained preoperatively from the parents of the hypospadias patients as part of the standard of care to be used as a pre-and post-operative clinical reference. The images showed the entire ventral shaft of the penis, with glans and meatal location. For all images, five ground truth anatomical landmarks – A, B, B', C and C' – were created and verified by two expert pediatric urologists according to the POST landmark definitions [15]. Briefly, A = distal midline mucocutaneous junction; B and B' = glanular knobs where the mucosal edges of the UP's edge change direction; C and C' = glanular/coronal junctions. Additionally, to ensure reliable representation of the target domain, the collected dataset featured a variety of backgrounds, lighting conditions, patient skin colors, camera angles, and UP qualities, along with other hypospadias variants such as megameatus intact prepuce. Due to the different specifications of the capturing cameras, these images were of variable size and quality. For that reason, a rigorous quality control process was used to ensure the quality of the dataset by identifying and removing images with extreme camera angles (e.g., lateral images where landmarks could not be recognized) or with extremely low quality. Supplementary Figure 2 shows sample images from the compiled dataset.

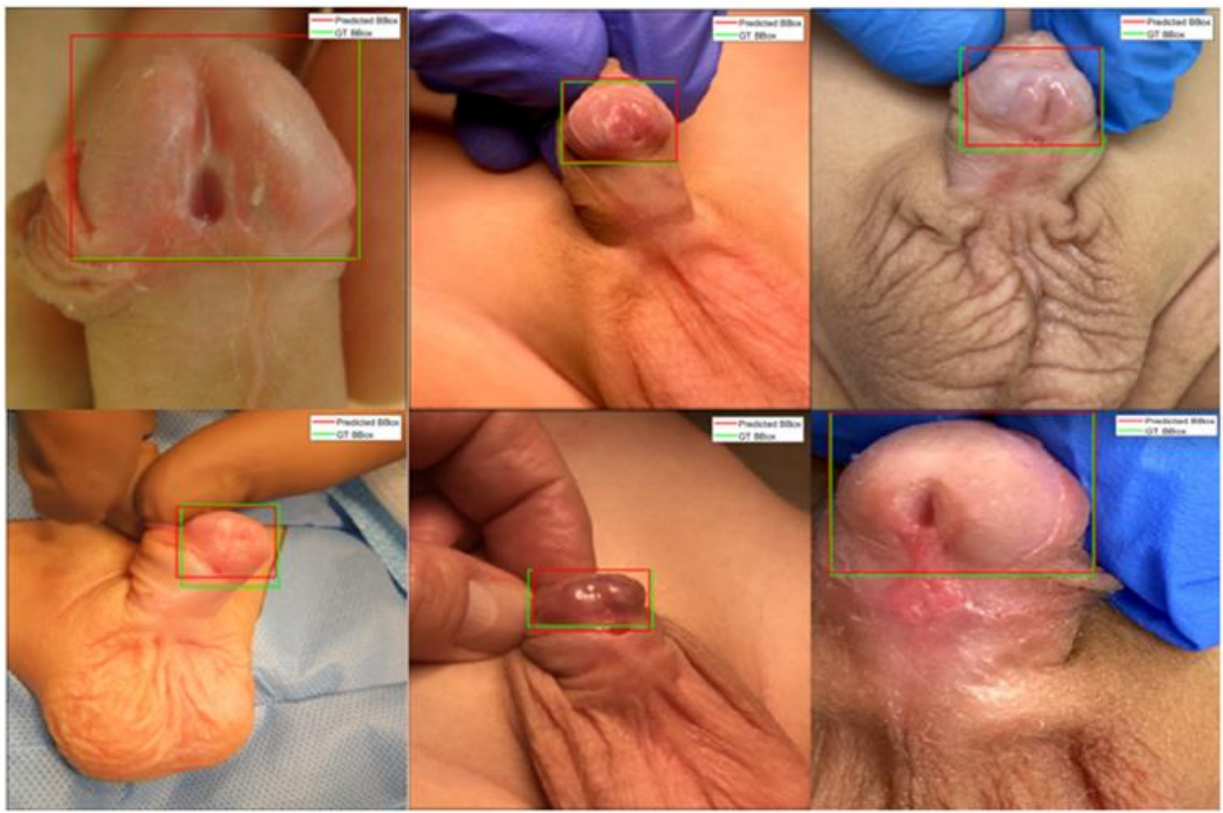

**Figure 2**. Sample qualitative evaluation of the glans localization model. Green bounding boxes are manually labeled ground truth, while red bounding boxes are model predictions.

### 2.1.1 Glans localization

Glans detection and localization form a solid foundation for automatic POST landmark detection. They were performed via a robust and accurate localization using bounding boxes to crop the glans area from the original input image, thereby eliminating redundant and irrelevant areas while preserving only the segments where the glans area was captured. This localization was performed using a YOLOv5 network [23], a recent update to the well-known YOLO (You Only Look Once) series. Similar to many other object detection networks, YOLOv5 relies on partitioning an input image into a grid of sub-regions. The predictions of the bounding boxes, along with their confidence probabilities, are then estimated per grid location. Four different variants of YOLOv5 – namely YOLOv5s, YOLOv5m, YOLOv5l, and YOLOv5x – have been described in the literature [24]. Despite their differences, these variants all share similar basic components. YOLOv5 architecture generally consists of three main parts: backbone, neck, and prediction head. Backbone typically comprises cascading CNN blocks that are known for their strong feature extraction capabilities. In the neck stage, the extracted feature maps are then reprocessed to ensure that these features are used to their full potential when predictions are made. Multiple upstream and downstream pathways, along with additional skip connections, are often used to assure high-resolution feature reusability. Finally, the prediction head is responsible for detecting the locations and the labels of the bounding boxes. A non-maximum suppression method is also used to eliminate overlapping predictions for the same target object [24]. It is worth noting that, while developing the glans localization network, we leveraged transfer learning by initializing the convolutional layer with pre-trained weights to ensure faster training and convergence.

### 2.1.2 POST landmark detection

High-Resolution Network (HRNet) is a recently described universal neural architecture that has achieved state-of-the-art performance in various computer vision tasks including human pose estimation, semantic segmentation, and object detection [25]. Unlike other common encoder-decoder type architectures, where the decoder network recovers high-resolution representations from low-resolution representations that the encoder part learns, HRNet maintains a high-resolution representation throughout the whole architectures. This results in a more spatially precise representation that is highly desirable, especially for position-sensitive tasks such as landmark detection. Concretely, this is achieved by parallel multi-resolution convolution streams, in which the network architectures start with a high-resolution convolution stream and then gradually add other high-to-low resolution streams. To ensure reliable feature fusion and reusability, information is regularly exchanged across parallel convolutional streams. This approach – concatenating convolution streams with different resolutions in parallel rather than in series – results in learning representations that are both semantically strong and spatially fine. Different schemes have been proposed for merging outputs with different resolutions at the prediction head stage. In this study, we used HRNetV2 [25], in which the outputs from the low-resolution convolution streams are unsampled and then concatenated with the higher-resolution feature maps. A 1×1 convolutional kernel is then used to produce the final activation map. In many cases, directly regressing the coordinates of the landmarks from the images may result in suboptimal localization performance. Heatmap regression is therefore a common approach in this type of problem [9]. In this technique, a 2-dimensional Gaussian is centered at the landmark position and used to create the ground truth heatmaps. Here, we set the standard deviation of the Gaussian to 1.5 (i.e. $\sigma_x = \sigma_y = 1.5$). Furthermore, mean squared error (MSE) loss was applied as a loss function during the training process.

### 2.1.3 POST score calculation

As mentioned previously, the images in the dataset had various aspect ratios and sizes. Because the ratio of pixel distances must be calculated during the last stage of the framework, resizing the images without preserving the original image aspect ratio would lead to a misleading calculation of the POST score. For that reason, the predicted landmarks were first mapped to the original image space, and the AB/BC ratio was then calculated for each of the individual pairs of points (left and right B/C pairs). Finally, the mean of the two ratios was reported as the final POST score.

### 2.2 Experimental setup

In all of the presented investigations, a 5-fold cross-validation scheme was used, while 20% of the training data were preserved for validating the models in addition to hyperparameter tuning and model selection. We used standard image augmentation techniques to expand the training dataset and to ensure higher robustness against images captured under different environments. These techniques included translation, ±30 degree in-plane rotation, $0.75 - 1.25$ scaling, and random flipping. To ensure a reliable and unbiased estimation of overall performance, we used the automatically cropped images from the glans localization network to develop and evaluate the landmark detection model.

In the glans area localization, stochastic gradient descent (SGD) optimization was used to train the YOLOv5s model for 250 epochs with an initial learning rate of $10^{-2}$ and a momentum of 0.937. A batch size of 16, and a weight decay of $5 \times 10^{-4}$ were also applied. In developing the landmark detection model, we used Adam optimization [26] with an initial learning rate of $10^{-4}$ and momentum updates of 0.9 and 0.99 for $\beta_1$ and $\beta_2$, respectively. The model was trained for 100 epochs with a mini-batch size of 16 images. An early stopping criterion with 15 epochs was employed to avoid overfitting, such that if the validation loss did not improve for 15 successive epochs, then the training would be terminated. A learning rate scheduler was implemented to drop the learning rate to $10^{-5}$ and $10^{-6}$ at the 50th and 70th epochs, respectively. Additionally, the PyTorch library [27] with Python 3.7 was used to train and evaluate the proposed pipeline, running on a PC with an Intel® Core™ i9-9900K CPU @ 3.60GHz and 32.0 GB RAM, with an 8 GB NVIDIA GeForce RTX 2080 SUPER GPU.

### 2.3 Evaluation metrics

#### 2.3.1 Object detection evaluation metric

The performance of the glans localization models was assessed using the mean average precision (mAP). The AP is defined as the area under the precision-recall curve, whereas the mAP is the mean of the AP over the classes:

$$mAP = \frac{1}{n} \sum_{i=1}^{n} AP_i \quad for\ n\ classes \qquad (1)$$

The mAP offers a solid foundation for evaluating the performance of the object detection models by comparing the ground-truth bounding box to the detected box.

#### 2.3.2 Landmark detection evaluation metric

To quantitatively evaluate the proposed landmark detection model, we employed the normalized mean error (NME), using the distance between the C points (i.e. the glanular diameter distance) as a normalizing distance to eliminate the effect of glans size:

$$NME = \frac{1}{n} \sum_{i=1}^{n} \frac{||S_i - \tilde{S}_i||_2}{d} \qquad (2)$$

In this equation, $S_i$, $\tilde{S}_i$ and $d_i$ represent the $i^{th}$ example ground truth landmark, the predicted landmark, and the glanular diameter distance, respectively. $|| S_i - \tilde{S}_i ||_2$ represents the Euclidian distance (i.e. the L2 norm).

Furthermore, we also reported the failure rate at a threshold of 0.1 ($FR_{0.1}$). This can be considered a measure of robustness against difficult examples.

## 3. RESULTS

This section describes the performance of the glans localization and the POST landmark detection models, along with a quantitative and qualitative evaluation of each stage.

### 3.1 Glans detection and localization

The glans detection model achieved an accurate and robust performance in localizing the glans area within the images, with a mAP of 99.5% and a sensitivity of 99.1%. Figure 2 presents a qualitative evaluation of the model performance. It is worth mentioning that YOLOv5s have presented a minor improvement in the detection performance when compared to other YOLOv5 variants.

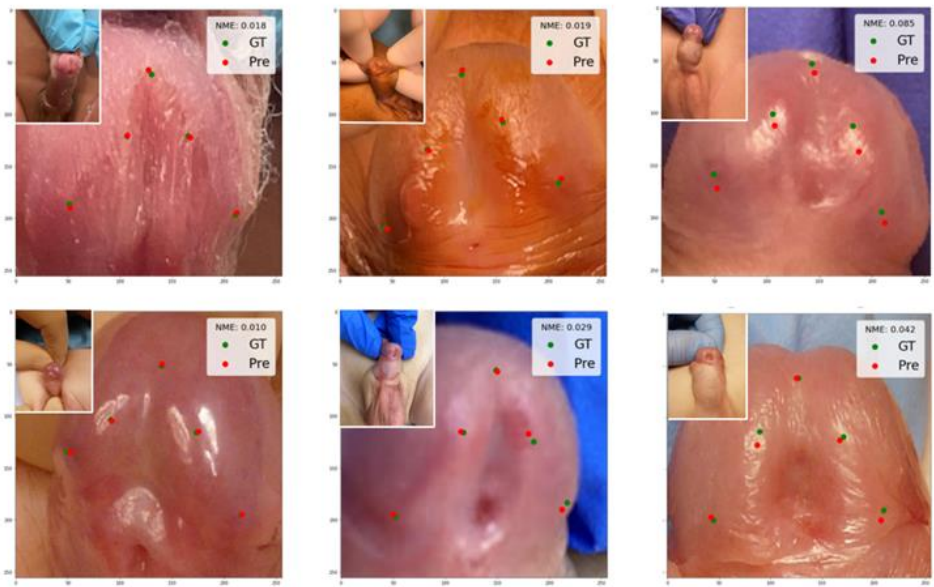

**Figure 3**. Sample qualitative evaluation of the estimated POST landmarks. Green points are manually annotated ground truth POST landmarks, whereas red points are model predictions landmarks.

## 3.2 POST landmark detection

The performance of the landmark detection models over the 5-fold test sets is illustrated in Table 1. As previously mentioned, NME is the main evaluation metric for the landmark detection model. The best performing model achieved an average NME of 0.07152 with a standard deviation of 0.004. Additionally, a 20.2% failure rate was achieved under a threshold of NME = 0.1. An MSE of 0.001 was also recorded, with a standard deviation of 0.0004. A qualitative evaluation of the landmark detection model is shown in Figure 3.

**Table 1**. Performance of POST landmark detection models for variable width. MSE = mean square error; NME = normalized mean error.

| Fixed parameters | Number of channels per Resolutions (Ordered from higher to lower) | NME | MSE loss |
|---|---|---|---|
| Representation head: HRNetV2 Heatmap size: 64×64 Number of stages: 4 Fusion method: Sum | (18,36,72,144) | 0.07152 | 0.001 |
| | (36,72,144,288) | 0.07481 | 0.001 |
| | (48,96,192,384) | 0.08216 | 0.002 |

It is clear from the presented results that the shallower model presented superior performance when compared to its wider counterparts. The tendency of the wider model to overfit the training data was inevitable as the number of channels per convolutions resolution path increases, even when common regularization techniques (e.g. weight decay and dropout) are employed.

### 3.3. Implementation

The previously described pipeline has been implemented, and a prototype system is publicly available for practitioners to use through a designated website. In this prototype system, a simple and straightforward registration process is required only during the first visit of the website. Once the account has been created successfully, the user will be prompted by general guidelines regarding the image capturing and uploading processes. Next, the uploaded image will be transferred to an AI-enabled server, where the necessary pre-processing and the previously described stages will be performed to determine UP quality. Supplementary Figure 2 illustrates the general framework of the implemented prototype system, which provides the user with a reliable and accurate POST score calculation along with the implied interpretations as suggestions. The front end of this system was implemented using the React library to ensure a user-friendly and responsive design. The back end was implemented in Flask and hosted on a Google Cloud virtual machine. The website developed for this study can be accessed through the following link: https://hypospadias-ai.netlify.app/.

## 4. COMMENT

Hypospadias literature on surgical management and outcome assessment has generally been based on subjective anatomical assessments of predictive variables (e.g., UP quality, meatus location, and ventral curvature) [14,28]. This subjectivity during the evaluation stage of most scoring tools can affect decision-making and communication among researchers when comparing inter-institutional outcomes [28].

Holland *et al.* examined the effect of UP quality, including UP depth, and correlated it to postoperative complications [11]. Such classification, however, is subjective and not reproducible. Likewise, Bhat's *et al.*'s definition was quite subjective, i.e., the UP was classified as wide, average, or narrow according to the surgeon's ability to tubularize the UP around a selected catheter [29].

The evaluation of the G component of the Glans-Urethral Meatus-Shaft score (GMS) is also highly subjective [30]. This component considers the UP quality in combination with assessment of the glans size, which ideally should be evaluated separately. On the other hand, Ru *et al.* studied the glans width (G) over the UP (U) and calculated the U/G ratio, and the measurements were captured under the erection test [31]. Another limitation of most studies is the arbitrary cutoff for UP measurement, considering 8 mm the landmark to assess outcomes [14]. Such selection is not applicable to various ages as well as variable penile sizes.

Although significant subjectivity was encountered in the literature on UP quality, several reports [11,32] showed better cosmetic outcomes in "favorable" UP, while Chukwubuike *et al*. found no statistical difference [33]. However, it is important to recognize that cosmetic evaluation is subjective and is often associated with significant disagreement among investigators.

The creation of an objective tool to classify hypospadias is therefore of critical importance. We have introduced the plate objective scoring tool (POST) as a reproducible and precise approach for quantifying UP quality and a guide to selecting particular surgical techniques based on that [15]. The POST score showed good inter- and intra-rater agreement when evaluating UP quality. However, inter-observer disagreement can exist when measuring the POST score from images that lack 3D spatial recognition, potentially leading to varying results.

There is an increasing interest within the pediatric urology community in a tool that standardizes the classification of hypospadias and its reporting in the literature. Our results support the possibility of further standardizing UP assessment from captured hypospadias images, and the use of machine learning algorithms and image recognition shows that these novel artificial intelligence technologies are useful for scoring hypospadias. The ability of the algorithm presented here to capture anatomical variables and measurements may lead to more objective evaluations and improve patient care and data collection for research purposes. A potential limitation of this study is the lack of measurements of different penile dimensions, since our available anonymous database did not include this information. Therefore, future studies incorporating different penile dimensions may be useful to expand the clinical utility of the algorithm shown in this paper.

## 5. CONCLUSIONS

In this paper, we proposed an automated approach for objective UP quality assessment. The implemented system was accurate and robust in glans localization and detection tasks, with a mAP of 99.5% and an overall sensitivity of 99.1%. Additionally, a deep high-resolution network architecture achieved superior performance in detecting the locations of the five POST landmarks, with an average NME of 0.07152. These results illustrate the potential benefits that the proposed framework can offer by increasing inter-rater consistency and standardizing UP quality scoring. Less experienced surgeons could benefit from real-time intraoperative application of this algorithm to aid decision-making and hence improve post-operative outcomes. Future applications of this technology may be used as a predictive tool for UP quality and surgical outcomes in hypospadias.

**Authors' statement**

**Ethical approval**

No ethical committee approval was required.

**Funding**

This research received a fund from Hamad Medical Corporation Medical Research Centre #20841.

**Competing interest**

All co-authors have nothing to declare.

**Supplementary Materials**

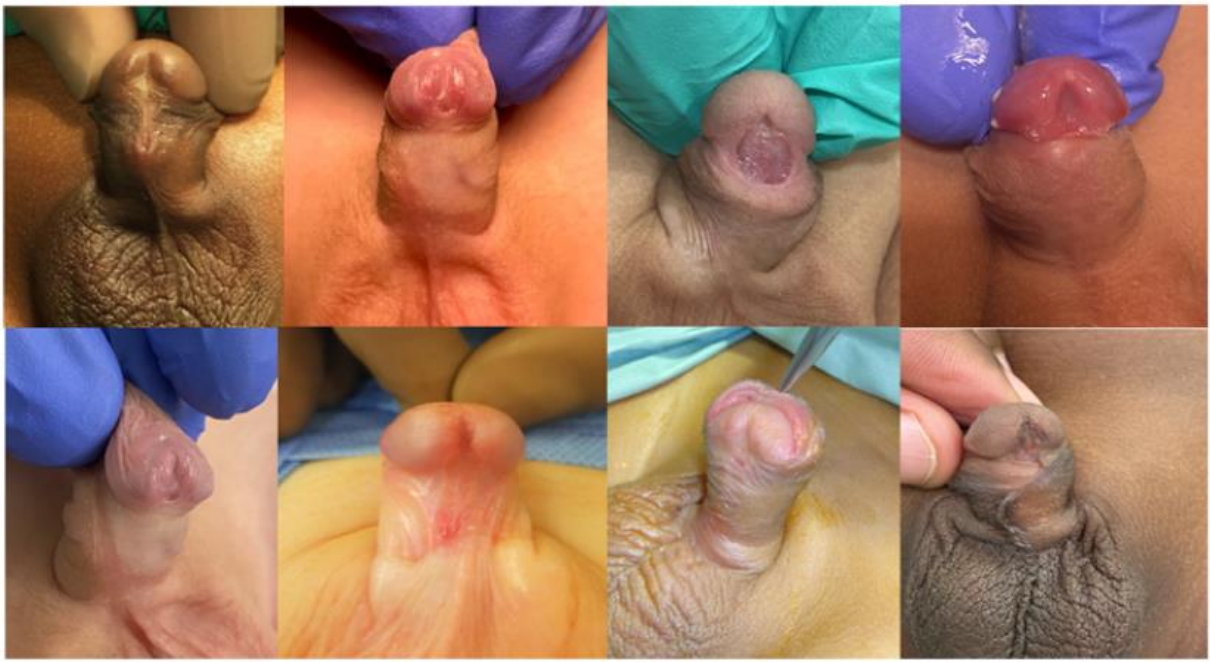

**Supplementary Figure 1**. Sample images from the compiled dataset, illustrating the range of image, patient, and hypospadias parameters included to ensure broad representation of the target domain.

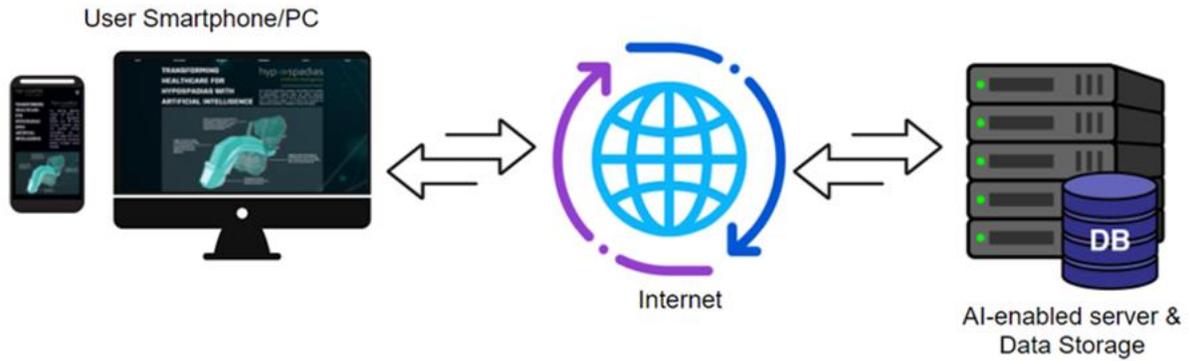

**Supplementary Figure 2**. Illustration of the general framework for automatic POST score calculation.